\documentclass[a4paper,twoside]{article}

\usepackage{epsfig}
\usepackage{subcaption}
\usepackage{calc}
\usepackage{amssymb}
\usepackage{amstext}
\usepackage{amsmath}
\usepackage{amsthm}
\usepackage{multicol}
\usepackage{pslatex}
\usepackage{apalike}
\usepackage{float}
\usepackage{listings}
\usepackage{xcolor}
\usepackage{svg}

\usepackage{SCITEPRESS}     

\begin{document}

\title{Convolutional Neural Network for Elderly Wandering Prediction in Indoor Scenarios}

\author{\authorname{Rafael F. C. Oliveira\sup{1}\orcidAuthor{0000-0002-0346-3073}, Fabio Barreto\sup{1}\orcidAuthor{0000-0003-4842-5238},Raphael Abreu\sup{1}\orcidAuthor{0000-0001-5917-6113}}
\affiliation{\sup{1}Centro Universitário Unilasalle do Rio de Janeiro, Niterói, Brazil}
\email{\{rafael.faustini,prof.fabio.barreto,prof.raphael.abreu\}@soulasalle.com.br}
}


\abstract{This work to propose a way to detect wandering activity of Alzheimer's patients from path data collected from non-intrusive indoor sensors around the house. Due to the lack of adequate data, we've manually generated a dataset of 220 paths using our own developed application. Wandering patterns in the literature are normally identified by visual features (such as loops or random movement), thus our dataset was transformed into images and augmented. Convolutional layers were used on the neural network model since they tend to have good results finding patterns specially on images. The Convolutional Neural Network model was trained with the generated data and achieved an f1 score (relation between precision and recall) of 75\%, recall of 60\%, and precision of 100\% on our 10 sample validation slice.}

\onecolumn \maketitle \normalsize \setcounter{footnote}{0} 

\section{\uppercase{Introduction}}
\label{sec:introduction}

\noindent 

Alzheimer is a type of dementia, characterized as a neurodegenerative disease that affects especially memory and behaviour. Wandering is one of the most common behavioural problems of Alzheimer's patients as seen in \cite{teri1988behavioral}.

Alzheimer's patients should be monitored at home to prevent accidents and moderate changes in the patient's behaviour, such as agitation, aggression and wandering behaviour \cite{ballard2009management}.
This task is usually done by family members or contracted professionals. In particular, the family caregiver is more susceptible to several pathologies due to the impact that this task has on the deprivation of time and its activities \cite{radziszewski2017designing}.

In the current context, the COVID-19 pandemic has increased even more the necessity to monitor Alzheimer patients, because they eventually suffer from anxiety and even increased health risks \cite{boutoleau2020impact}. Which in turn, contributes to the caregiver burden.

Since wandering behaviour is most times coupled with agitation in Alzheimer patients \cite{ballard2009management}, a system that automatically and correctly detects wandering behaviour in a smart-home can take actions to alleviate the agitation of the elderly, or trigger events to inform the caregiver that the elderly is wandering.  There are a variety of interventions to reduce agitation in elderly people with Alzheimer. Some of them can be automated inside a smart-home scenario. Such as aroma, music and lights \cite{o2011systematic}. Thus, by utilizing a smart-home to perform such tasks, a system can alleviate some of the caregiver burden. 

Therefore a prerequisite to such a system is to correctly detect a wandering behaviour. This work proposes a machine learning technique for identifying wandering patterns. The focus of the paper being the detection of wandering in an indoor scenario.

By observing the difficulties faced by caregivers in monitoring Alzheimer patients, the necessity of continuous supervision was noticed and it highlights the importance of this work. By using the Internet Of Things (IoT) \cite{karimi2013internet}, we could obtain the movement data of the elderly in a non-intrusive way, however, due to the massive quantity of data, it's hard to identify wandering patterns.

\begin{figure}[h]
     \centering
     \begin{subfigure}{0.4\textwidth}
         \includegraphics[width=6cm]{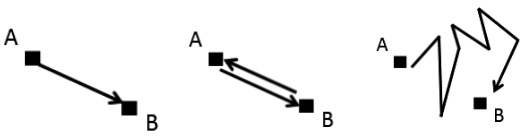}
         \caption{Direct, pacing and random} \hspace{1em}
     \end{subfigure}
     \hfill
     \begin{subfigure}[b]{0.4\textwidth}
         \includegraphics[width=6cm]{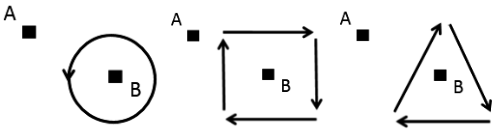}
         \caption{Lapping} \hspace{1em}
     \end{subfigure}
     \caption{Wandering and non-wandering movement patterns \cite{delaunay2017wandering}}
     \label{fig:delaunay}
     \hfill
\end{figure}

In Figure~\ref{fig:delaunay} we have 6 different patterns of movements between point A and a point B. It was identified that some of these patterns were more present in wandering than normal activity \cite{martino1991travel}, \cite{delaunay2017wandering}.
In Figure~\ref{fig:delaunay} (a) the first pattern is present in both wandering and non-wandering movements, but mostly in non-wandering movement. So it was not associated as a wandering activity, differently from the other two patterns of this figure, pacing, and random. Figure \ref{fig:delaunay} (b) shows the most likely patterns of wandering movements called lapping, where the movement turns on a loop around a point. 

Due to the lack of data of normal and wandering movement in the same scenario, we've reproduced manually the likely wandering patterns represented in Figure~\ref{fig:delaunay} to create the anomalous activity data and also normal data mostly composed of direct movements. 

Thus, in this work is proposed a method that can based on the estimated recent position with IoT and a neural network model for the deep learning approach, infer the patient stress so that we can act in trying to mitigate the negative effects of it. It is hoped that this can be an important tool to help caregivers and relatives of Alzheimer's patients. 

\begin{figure}
    \centering
    \includegraphics[width=7cm]{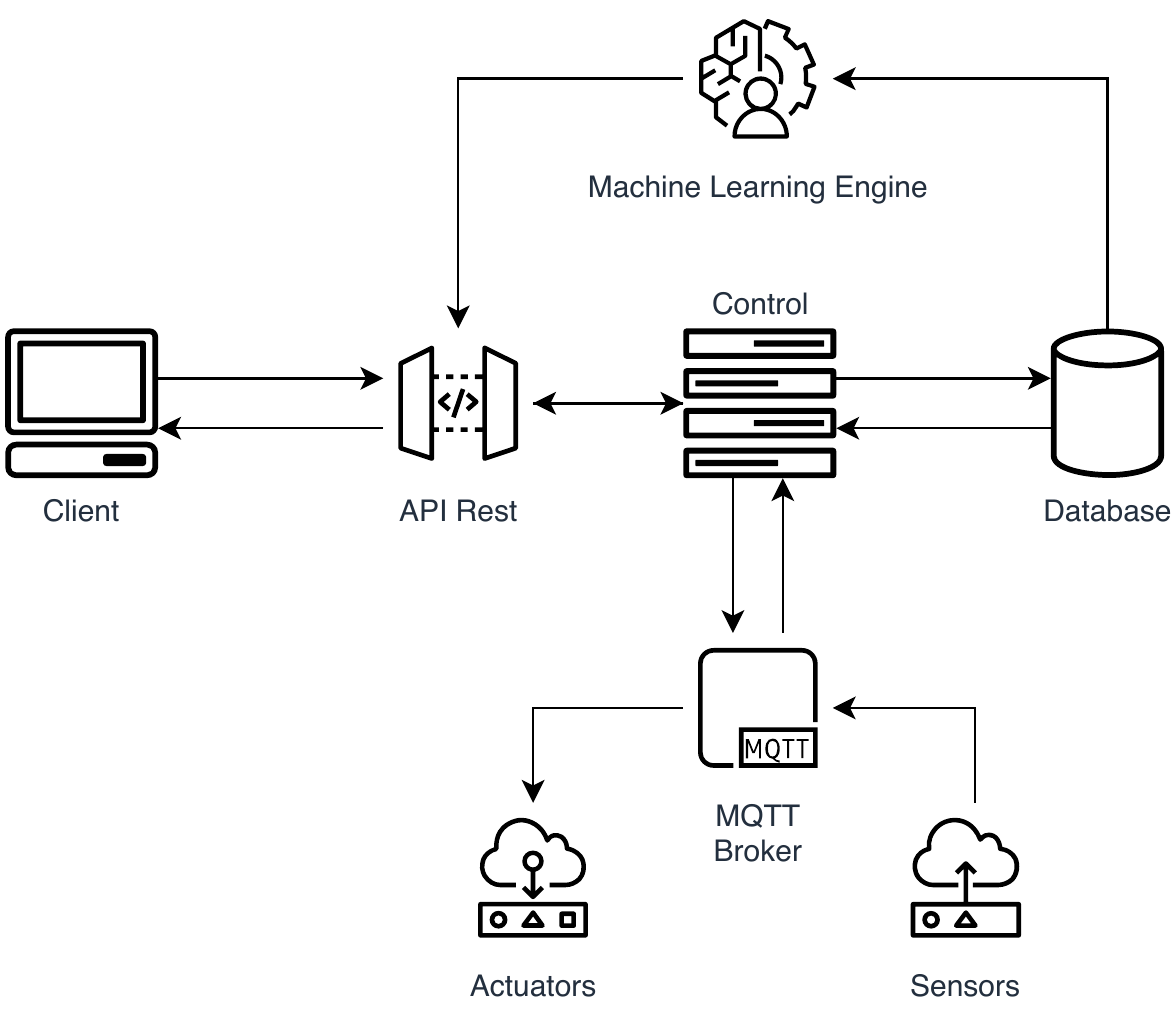}
    \caption{Architecture of the system}
    \label{fig:smartcarearchitecture}
\end{figure}

Even though this work reaches especially the machine learning part of the project, to understand it is crucial to have an overview of the entire system's architecture. The system's architecture represented in Figure \ref{fig:smartcarearchitecture} is divided into seven components, each one is part of the flow from the client to data capturing. The Client is the interface of access and system's configuration, REST API \cite{masse2011rest}  is a double way to communicate with other components. With that, the client can request to the API, data from the machine learning engine and NoSQL Database \cite{han2011survey}.
Temporal data from the database is analyzed and used to create a predictive model that can infer stress activities in Machine Learning Engine, which is the one that was approached in more depth by this work. Treatment of the sensors data, data fusion between sensors, and intermediate system communications are the roles of the controller. The controller treats the data, by analyzing the raw simple sensors activations around the house and, with data aggregation, generates the path of the elderly that will be used to identify anomalies. Raw and treated data are stored in the Database. The data is captured through non-intrusive sensors such as distance, presence, pressure sensors. It is the sensor part of the IoT module, in which direct communication protocol between devices is done with the Message Queuing Telemetry Transport(MQTT) interface \cite{light2017mosquitto}. Also in the IoT module, we have the actuators which are responsible to help to mitigate the elderly stress, being able to produce aromas, and turn on and off the lights. The work shown in this paper is based on this entire architecture. With this, our work presents a method to utilize machine learning to detect wandering using data coming only from non-intrusive sensors.

This work is organized as follows, the related work of IoT monitoring and the use of machine learning will be shown in Section~\ref{sec:relatedwork}. Section~\ref{sec:dataset} overviews the application developed for the data generation and the composition of the data used in our learning structure. Section~\ref{sec:implementation} approaches the data pre-processing and the construction of our neural network model. In Section~\ref{sec:results} we discuss the performance of our model and experiments. Finally, in Section~\ref{sec:conclusion} we show future applications and conclude our work.

\section{\uppercase{Related Work}}
\label{sec:relatedwork}

\noindent As stated in \cite{ordonez2016deep}, human activity recognition (HAR) is based on the assumption that specific body movements translate into characteristic sensor signal patterns, which can be sensed and classified using machine learning techniques. The use of machine learning was already present in activity tracking, an example is \cite{akbar2019smart} that uses data from wearable sensors such as a wristband for heart rate monitoring and neural networks to learn sleep patterns of the users, in order to turn off the lights after they fall asleep. And as well in \cite{oniga2014human}, IoT data is used to train an artificial neural network model that, given the information of the arm and body pose of the patient, it could track activities such as the walking movement or even predict if the user was sitting down.

\cite{khan2018detecting} surveys and compare the literature about agitation and aggression detection. The tracking modalities presented was by using wearable devices, computer vision and multimodal, it also presents a correlation between agitation and aggression in dementia patients and actigraphy. Relating this overview of papers to our work, we focus on the well-being of the patient by assuming that the ambient only employs sensor-tracking. That the elderly do not need to wear, neither feel the hostility of a camera watching him.

In \cite{vogado2020leuknet} CNN (Convolutional neural network) is used to help the diagnostic of leukemia, using images of blood smears, complemented with the application of transfer learning and data augmentation techniques, reaching a result of 98.28\% of accuracy. Compared to our work, we use CNNs that can detect anomalies in images and also the use of data augmentation to generate more data.

In \cite{amin2020convolutional} and \cite{amin2020brain}, machine learning is used to detect brain tumors. \cite{amin2020convolutional} uses CNNs and \cite{amin2020brain} autoencoders to approach the problem. They were trained and evaluated using BRATS challenges databases and had good results in comparison to the literature. In relation to our work, we similarly have an image classification problem. However, in the work of \cite{amin2020convolutional}, the CNN has 14 layers whereas our networks have only 4 layers.

\cite{doughty1998diana} proposes a system to monitor patients with dementia using non-intrusive sensors to track their activities. The way used to detect wandering was by counting event activations such as sitting down, moving to the chair, opening and closing drawers, and then calculating a score based on repeated events 5 minutes. The advantage of the Machine Learning approach, proposed in our work, is the easy adaptability to the patient. Our method is based not only on repetitions but on any data pattern in the movement. With the advantage of adapting to new patterns if needed.  

Contrasting to related work, our method focus on predicting wandering using sensing information collected in a non-intrusive way. Similar to \cite{doughty1998diana}, but coupled with a machine learning approach which makes possible the detection of known patterns in the literature such as \cite{martino1991travel}. Also, being able to detect other patterns by the data distribution with the machine learning engine and having a lightweight architecture, capable of running on an IoT device.



\section{Dataset} \label{sec:dataset}

\noindent Due to the lack of indoor datasets containing normal and anomalous activity, the data generation strategy was adopted similar to the one presented in \cite{arifoglu2017activity}. Therefore, an application to simulate these paths was developed. With our tool, the user can manually insert, annotate and visualize movement data point-by-point, without the need of a real patient. With this tool, we generated movement patterns of wandering and normal activity.

\begin{figure*}[ht]
    \centering
    \includegraphics[trim={0 7cm 0 0},clip,width=15.5cm]{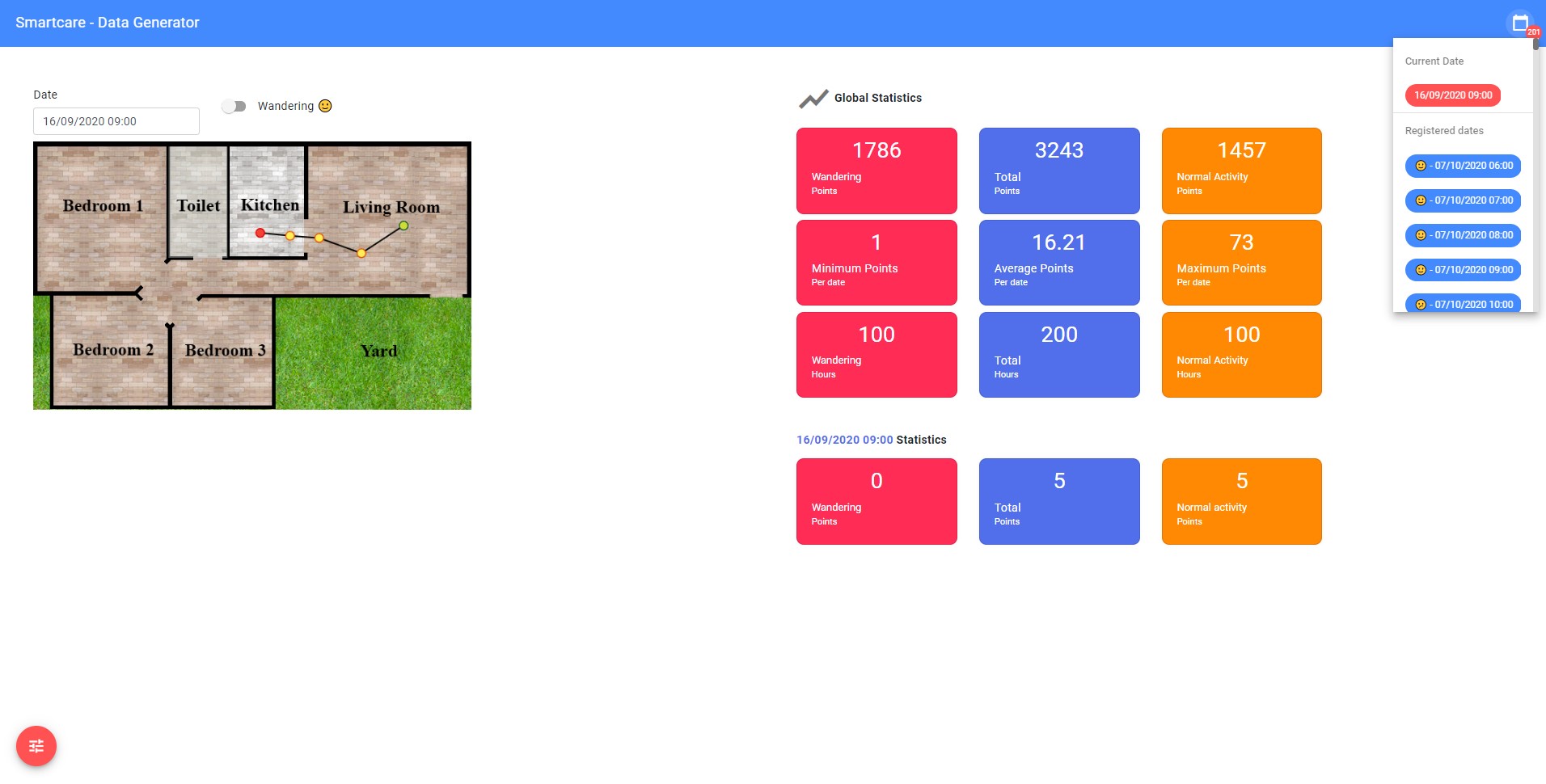}
    \caption{Developed application for data generation. The user interacts with the floor-plant to place movement points, define the hour of the movement and if it was wandering. 
    On the right side of the page shows the statistics created to visualize the generated data. In the upper right a menu that shows all the registered movement dates and hour. }
    \label{fig:geradordados}
\end{figure*}

In order to make easier the work of data generation, we've created a web application, as shown in Figure~\ref{fig:geradordados}.  In the application, we can upload a floor plan image and set the points in the ground floor plan. Every point represents a sensor reading indicating the position of the patient. In this application we defined the interval of analysis of every path created to be hourly, since a path is a group of points with this delimiter, we can classify the entire hour path as wandering or normal activity.  Is important to note that although the generated data of the datasets were synthetic and don't represent the real world, they were generated to simulate the real-world scenario. Since the architecture of the neural network is simple and robust, it is expected that it generalizes in the real world with real data in a similar manner as the work of \cite{arifoglu2017activity} does.

As seen in Figure~\ref{fig:geradordados}, on the left we have the ground floor plan and on the right the statistics created to help to monitor the data balance, such as the number of hours with wandering and normal activity. The visual part of the application was developed using Vue.js, Vue Material, and bootstrap libraries. The ground floor plan and the points mechanism P5.js library was used, which simplifies the JavaScript canvas. The application is available at https://github.com/Unilasalle-SmartCare/smartcare-frontend.

The created data using our application was exported in JavaScript Object Notation (JSON) \cite{severance2012discovering} format. The file structure is based on an array of objects, where each object represents a location point in the environment, each point consists of four attributes. 'x' and 'y' represents it's respective coordinates in the ground floor plan image starting from the upper left. The 'date' attribute, which date and hour the point belongs. Each point has the attribute stress that indicates if the elderly is wandering. Even though each point has the attribute stress, the application considers a time interval to account for a wandering activity. It is not able to have two points in the same time interval with different stress values. In order to know where the movement starts and ends, the points of the same data and interval are inserted in the order as the elderly movement.

With the application, we developed two datasets. The first is a train/test dataset and the second is a validation dataset. The generated train/test dataset consists of a total of 200 hours, being 100 hours of normal activity and 100 hours of stress. As seen in table \ref{tab:datasettrain}, they are sliced into training and test part being respectively 75\% and 25\%. The validation dataset to be used in the evaluation of our machine learning model is consisted of 20 hours, being 50\% anomalous activity and the other 50\% normal activity. The composition can be seen in more depth in table \ref{tab:datasettest}.

For the purpose of generating consistent anomalous data, patterns present in \cite{martino1991travel}, \cite{delaunay2017wandering} were reproduced manually using our developed data generation application.  We assumed normal movement as being not anomalous or random. The patterns classified as anomalous are Lapping and Pacing, seen in \cite{martino1991travel}. Random patterns are described by \cite{martino1991travel} as roundabout or haphazard travel to many locations within an area without repetition and is mostly composed of direct movements.

\begin{table}[h!]
\caption{Distribution of the train/test dataset}\label{tab:datasettrain} \centering
\begin{tabular}{|c|c|}
  \hline
  Movement Type & \# Samples \\
  \hline
  Lapping & 59 \\
  \hline
  Random & 11 \\
  \hline
  Pacing & 30 \\
  \hline
  Normal & 100 \\
  \hline
\end{tabular}
\end{table}

\begin{table}[h!]
\caption{Distribution of the validation dataset}
\label{tab:datasettest} \centering
\begin{tabular}{|c|c|}
  \hline
  Movement Type & \# Samples \\
  \hline
  Lapping & 7 \\
  \hline
  Random & 2 \\
  \hline
  Pacing & 1 \\
  \hline
   Normal & 10 \\
  \hline
\end{tabular}
\end{table}

\section{Implementation} \label{sec:implementation}

\begin{figure*}[h!]
    \centering
    \includegraphics[width=15cm]{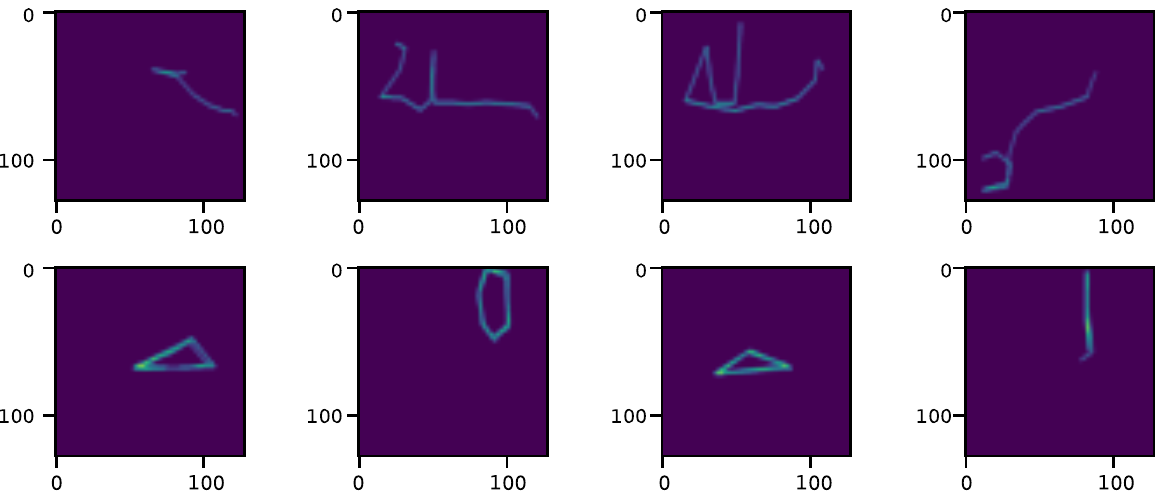}
    \caption{Samples of the path data plotted as images after rescaling}
    \label{fig:datasamples}
\end{figure*}

\textbf{Data Processing}. With the generated dataset, it's still needed to do some processing in order to use our artificial neural network model (ANN). First, all the points were grouped by their date and hour. Then, the points were converted to grayscale images. To create such images,  each point is converted and placed in the corresponding location on the image and a straight line is plotted between two adjacent points. As seen in Figure~\ref{fig:datasamples}, the resulting image shows a stroke where the patient moved in each hour.

Each pixel of an image consists of a value between 0 and 255. To make the values smaller and faster for our model, we normalized the pixels to be between 0 and 1, maintaining the distribution and ratio of the data. Since the images have the resolution of the ground floor plan, it would be a waste of resources to give to the model big images with very small features and it would impair the generalization of the model. To account for the aforementioned problems, the images were rescaled to 128x128 resolution.

\begin{figure*}[h]
    \centering
    \includegraphics[width=12cm]{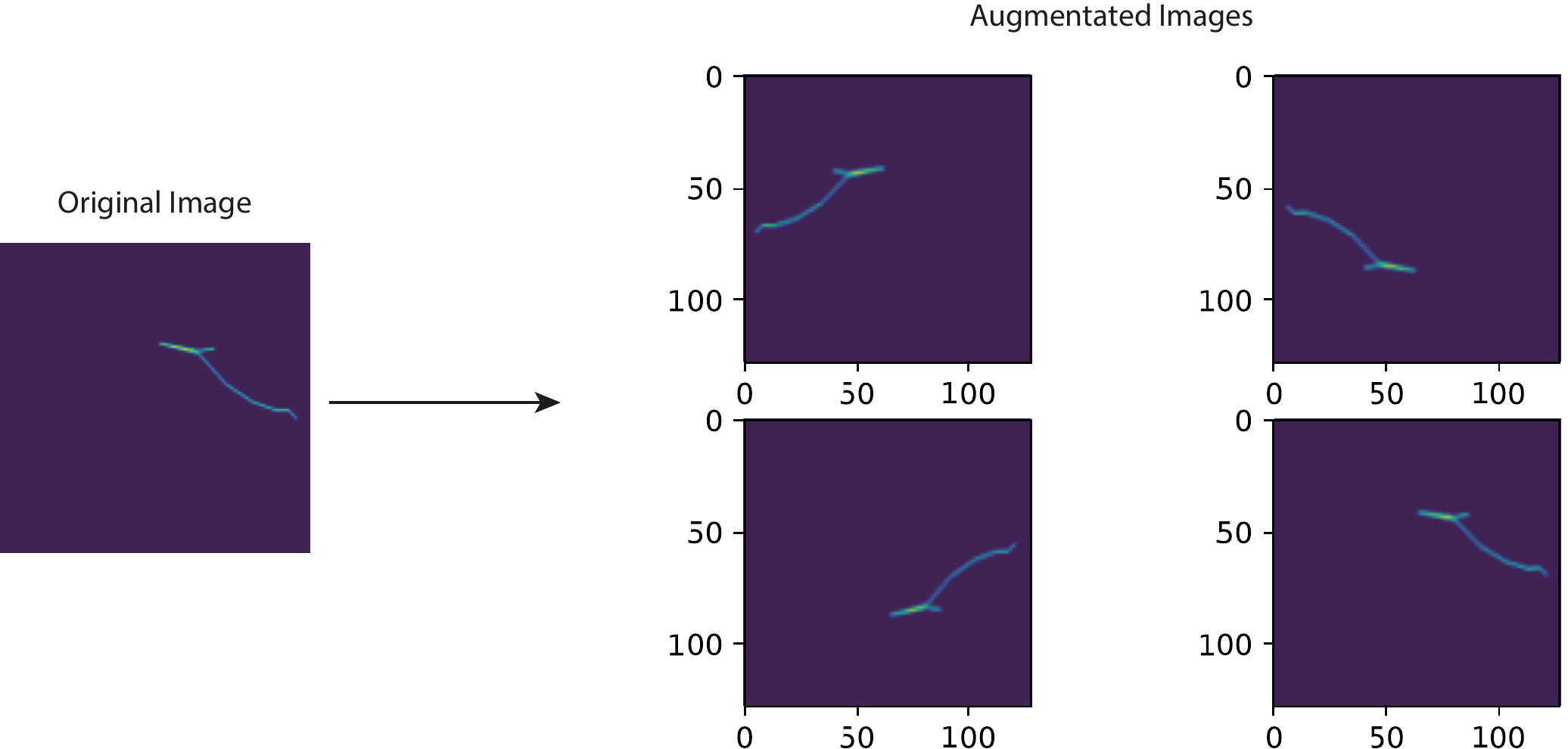}
    \caption{Data augmentation technique applied in movement path}
    \label{fig:dataaugmentation}
\end{figure*}

\textbf{Data Augmentation}. Even with the generated dataset we do not have much data. In order to prevent overfitting and help the model generalization, we used data augmentation techniques. As seen in Figure \ref{fig:dataaugmentation}, each train image generates more training images. 

Differently from \cite{vogado2020leuknet} where zoom and shearing were applied to the new images, we only rotated each image to a maximum interval of 3 degrees and randomly flipped vertically and horizontally.

\begin{figure*}[ht]
    \centering
    \includegraphics[width=14cm]{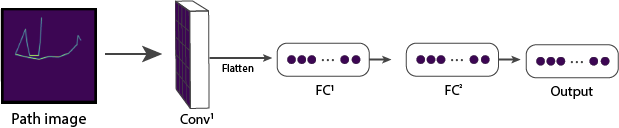}
    \caption{Architecture of our neural network model.}
    \label{fig:neuralnetworksmodel}
\end{figure*}

\textbf{Artificial Neural Network Model}. By using path images, it was seen that the Convolutional Neural Network (CNN) approach could be an interesting idea. Since it could identify the anomalous movement activities by learning the shapes and pixel distribution of the paths. The CNN architecture is presented in Figure~\ref{fig:neuralnetworksmodel}. In the Figure, the processed image is given as input and goes into a convolutional layer ($conv^1$) with a filter size of 32, filtering is used to highlight features of the input. Max pooling is applied to reduce the dimensionality and make the neural network training faster, and then the output is flattened into an array. After that, we have three fully connected layers ($FC^1$,$FC^2$, and $FC^3$). Being the last one the output of our network between 0 and 1, which represents whether the input path is predicted as wandering or not. To avoid overfitting, we randomly dropped 25\% of the hidden units between the fully connected layers ($FC^1$ and $FC^2$) \cite{srivastava2014dropout}.

For the prediction, it is used the backpropagation algorithm that on training, a process where the network is fed with training data to learn. The algorithm constantly passes on each layer and by calculating predictions and measuring errors, it adjusts the the weights of the neurons to reduce errors \cite{10.5555/3153997}. From the convolutional to the fully connected layers, we opted to use Rectifier activation function (ReLU) in all of them since it reduces the probability of vanishing gradient problem \cite{talathi2015improving}. The loss function was binary cross-entropy that is good with binary classification problems.

The architecture was implemented using Keras \cite{chollet2015keras} with TensorFlow \cite{tensorflow2015-whitepaper} as backend.
To train the network, we used 150 epochs and trained in mini batches of 64 samples with a learning rate of 0.001.
The model and dataset will be available at https://github.com/Unilasalle-SmartCare/smartcare-machinelearning.

\section{Results}

\begin{figure}[H]
    \centering
    \includegraphics[width=7cm]{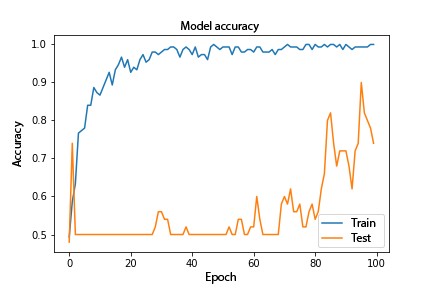}
    \caption{Accuracy per epoch of the model}
    \label{fig:graficotreinamento1}
\end{figure}

\begin{figure}[h!]
    \centering
    \includegraphics[width=7cm]{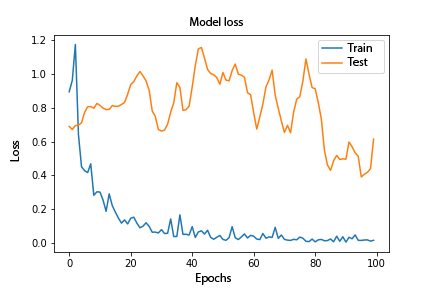}
    \caption{Learning curve representation of the model}
    \label{fig:graficotreinamento2}
\end{figure}

\noindent The accuracy of the model, which represents the model performance on the train/test data, is presented Figure~\ref{fig:graficotreinamento1}. The loss, that is the distance between the true target and the prediction, is presented in Figure~\ref{fig:graficotreinamento2}. The model was trained for 100 epochs with the train/test dataset. The optimizer algorithm used was the Adaptive Moment Estimation (adam).

The tuning of the model hyperparameters based on the train/test set and evaluating on the same set would leak information of the validation set on the model that may influence the results \cite{10.5555/3203489}. Based on that, besides the test set used in the training to calculate the model accuracy and loss seen in Figures~\ref{fig:graficotreinamento1} and \ref{fig:graficotreinamento2}, we've generated separately a validation dataset, consisted of 10 hours of stress and 10 hours of normal activity, an equivalent of 10\% the size of the data used for training with the objective to evaluate our model and avoid information leaks.

To measure accuracy of the models, we utilized the following evaluation metrics: precision, recall and f1 score. In Equations 1,2 and 3 we present each one. 

\begin{equation}
Precision = \frac{true positives}{true positives + false positives}
\end{equation}
\begin{equation}
Recall =  \frac{true positives}{true positives + false negatives}
\end{equation}
\begin{equation}
F1 = 2\times\frac{precision \times recall}{precision+recall}
\end{equation}

With the validation dataset, we could evaluate our final model and had a precision of 100\% followed by a recall of 60\% and the f1 score of 75\%.
The hyperparameters were chosen empirically by adjusting based on the two curves.

\begin{figure}[h]
    \centering
    \includegraphics[width=5cm]{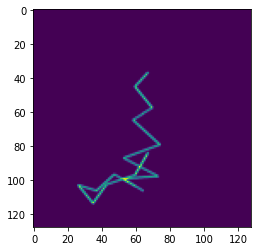}
    \caption{Sample of a random path generated with wandering activity}
    \label{fig:randomprediction}
\end{figure}

The trained model was able to identify all the movement patterns presented in Table \ref{tab:datasettest}. Figure~\ref{fig:randomprediction} shows a sample of random path generated from the validation dataset that was identified by our model correctly as wandering. 

The model obtained a satisfactory result considering the small amount of data. Besides that, the proposed architecture is lightweight and can be embedded into small scale devices. Therefore it is expected that the machine learning approach proposed in this work can predict not only patterns as the ones present in the literature \cite{martino1991travel}, \cite{delaunay2017wandering}, \cite{lin2012detecting} but also non-perceived ones. As raised by \cite{lin2014managing} it is important to perform a personalized user wandering recognition model, since those wandering activities might have different characteristics depending on the person and even environmental conditions. Our method paves the way for a reinforcement learning approach that can be used for the network to adapt itself to the specific elderly movement in that house.

\label{sec:results}

\section{\uppercase{Conclusions}}
\label{sec:conclusion}

\noindent In this paper, data preparation and lightweight CNN architecture were proposed to detect the wandering of Alzheimer's patients using data from non-intrusive sensors. By transforming all the data paths into images, we could use techniques such as data augmentation to expand the simulated dataset and reach better results with a convolutional neural network. 
Conventional tracking such as wearables and cameras enters into a deep discussion about the privacy of the users. Therefore, one of the challenges of this work was to propose a predictive method that could fit the environment with non-intrusive sensors. Another big challenge was the lack of data to develop the machine learning model, so we've generated using known patterns to simulate the real world, although the data being synthetic, it is expected that the machine learning architecture also works with real-world data.

For future work, collecting real patient's data is vital to validate our model in the real world and also learn even more wandering patterns. The conversion from the path to images as done in this work, can be expanded to other disease movement anomalies and also outside the health field. For example, the same method can be used from classifying potential customers of a store based on the movement. This comes with as much needed privacy concern, as the only input our method needs is a path of movement inside an indoor scenario. In the future, this idea can also be applied to identify wandering in an outdoor scenario. The use of reinforcement learning is a future challenge to bring interesting comparative results with this work. Since reinforcement learning tends to choose actions to maximize a reward based on the environment information \cite{10.5555/3203489}, these actions could be something to mitigate the patient stress such as aromatherapy or music.

\section*{\uppercase{Acknowledgements}}
\noindent The authors would like to thank Unilasalle-RJ for encouraging and financially supporting this work.

\bibliographystyle{apalike}
{\small
\bibliography{references}}

\end{document}